\documentclass{tlp}

\usepackage{amsmath}
\usepackage{graphicx}
\usepackage{multirow}
\usepackage{amsmath}
\usepackage{booktabs}%
\usepackage{xcolor} %
\usepackage{subcaption} %
\usepackage{amsfonts} %
\usepackage{nicefrac,xfrac}
\usepackage{verbatim} %
\usepackage{pdfcomment}

\usepackage[]{todonotes}
\usepackage[ruled,noend,linesnumbered]{algorithm2e}%

\newcommand{\asp}{\textsc{ASP}}
\newcommand{\goal}[1]{G{#1}}

\newcommand{\clingo}{\texttt{clingo}}
\newcommand{\clingraph}{\texttt{clingraph}}
\newcommand{\hc}{\textsc{HC}}
\newcommand{\np}{\textsc{NP}}
\newcommand{\fnp}{\textsc{FNP}}

\newcommand{\mypath}{\textit{p}}
\newcommand{\hs}{\textsf{HS}}
\newcommand{\mhs}{\textsf{MHS}}
\newcommand{\ms}{multi-shot}
\newcommand{\MS}{Multi-Shot}

\newcommand{\gs}{\textsf{General Snake}}

\newcommand{\approach}[1]{\textsf{#1}}
\newcommand{\os}{one-shot}

\newcommand{\aos}{\approach{one-shot}}
\newcommand{\Aos}{\approach{One-Shot}}
\newcommand{\preground}{\approach{preground}}
\newcommand{\Preground}{\approach{Preground}}
\newcommand{\redo}{\approach{ad hoc}}
\newcommand{\Redo}{\approach{Ad Hoc}}
\newcommand{\nogood}{\approach{nogood}}
\newcommand{\Nogood}{\approach{Nogood}}
\newcommand{\assume}{\approach{assume}}
\newcommand{\Assume}{\approach{Assume}}
\newcommand{\snakes}{\approach{snake}}
\newcommand{\Snakes}{\approach{snake}}
\newcommand{\bl}{\texttt{\char`\{}}
\newcommand{\br}{\texttt{\char`\}}}
\newcommand{\E}[1]{\textsf{E#1}}

\newcommand{\figsize}{0.285\textwidth}

\newcommand{\apple}{\pdftooltip{\includegraphics[height=0.6\baselineskip]{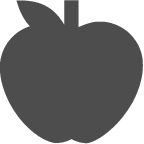}}{apple symbol}}
\newcommand{\snake}{\pdftooltip{\includegraphics[height=0.6\baselineskip]{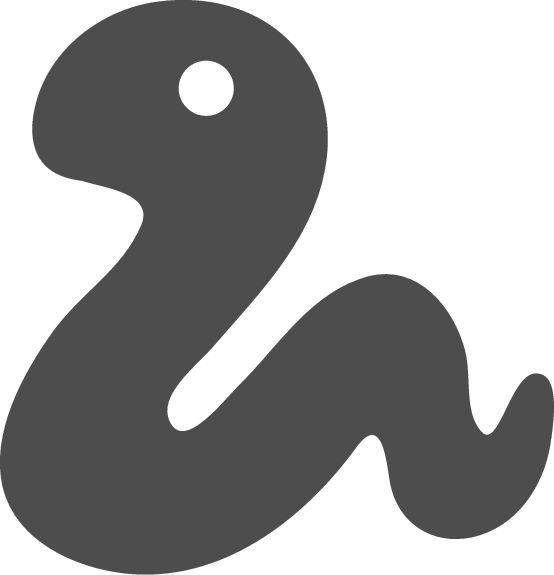}}{snake symbol}}
\newcommand{\minisnake}{\pdftooltip{\includegraphics[height=0.35\baselineskip]{img/snake.eps}}{snake symbol in index size}}

\DeclareCaptionSubType*{figure}
\SetArgSty{textnormal} %

\usepackage{xspace}

\newcommand{\cvar}[1]{\textcolor[RGB]{49, 132, 149}{#1}}
\newcommand{\cmat}[1]{\textcolor[RGB]{84, 98, 115}{#1}}
\newcommand{\cnum}[1]{\textcolor[RGB]{0, 0, 205}{#1}}

\begin{document}
	\SetKwComment{Comment}{// }{} %
	\DontPrintSemicolon	
	\SetKwIF{If}{ElseIf}{Else}{if}{$\!$:}{else if}{else \!:}{end if}%
	\SetKwFor{For}{for}{$\!$:}{}
	\SetKwFor{While}{while}{$\!$:}{}
	\SetKwRepeat{Do}{do \!:}{while}%

\lefttitle{Cambridge Author}

\jnlPage{1}{8}
\jnlDoiYr{2021}
\doival{10.1017/xxxxx}

\title[Winning Snake with \MS{} ASP]{Winning Snake: Design Choices in \MS{} ASP}

\begin{authgrp}
\author{\gn{Elisa} \sn{Böhl}} %
\affiliation{Logic Programming and Argumentation, TU Dresden, Germany}
\author{\gn{Stefan} \sn{Ellmauthaler}} %
\affiliation{Knowledge-Based Systems Group, ScaDS.AI / cfaed, TU Dresden, Germany}
\author{\gn{Sarah} \gn{Alice} \sn{Gaggl}}
\affiliation{Logic Programming and Argumentation, TU Dresden, Germany}

\end{authgrp}

\history{\sub{xx xx xx;} \rev{xx xx xx;} \acc{xx xx xx}}

\maketitle

\begin{abstract}
Answer set programming is a well-understood and established problem-solving and knowledge representation paradigm.
It has become more prominent amongst a wider audience due to its multiple applications in science and industry.
The constant development of advanced programming and modeling techniques extends the toolset for developers and users regularly.
This paper compiles and demonstrates different techniques to reuse logic program parts (\ms{}) by solving the arcade game \snakes{}.
This game is particularly interesting because a victory can be assured by solving the \np{}-hard problem of Hamiltonian Cycles.
We will demonstrate five hands-on implementations in \clingo{} and compare their performance in an empirical evaluation.
In addition, our implementation utilizes \clingraph{} to generate a simple yet informative image representation of the game's progress.

\end{abstract}


\section{Introduction}

Answer set programming (\asp{})~\citep{brewka2011answer} is an established declarative programming paradigm based on stable model semantics~\citep{gelfond1988stable}. 
It combines attributes of different fields of computer science such as knowledge representation, logic programming and SAT solving 
and is suited for numerous tasks like planning and configuration problems \citep{mainpotassco}. 
\asp{} is of interest as theoretical concept for computer science. 
Its declarative way of describing problems in a rule-based language allows for fast prototyping, offering great value for product applications as well. 
The ability to solve hard, discrete optimization problems makes \asp{} appealing for industrial applications. 
In  recent years, different specialized \asp{}-tools such as \clingo{}~\citep{PotasscoUserGuide19}, \texttt{DLV}~\citep{adrian2018asp}, \texttt{WASP}~\citep{alviano2013wasp}, or \texttt{alpha}~\citep{weinzierl2017blending} where introduced.
\clingo{} established itself as prominent choice, with its various tools and in-depth documentation, continuous development and many publications on  various problems \citep{gaggl2015improved,takeuchi2023solving,rajaratnam2023solving}.

With the growth of its applications, the implementation of advanced logic programs gains traction. 
Traditionally, the \clingo{} workflow grounds a logic program (i.e. replacing variables by contents) and solves the resulting program afterwards. 
In an iterative setting, this step is repeated for a similar, slightly different logic program.
This opens an opportunity to alter and re-solve the same logic program to reduce grounding and solving overhead. Especially grounding often states a significant bottleneck~\citep{BesinHW23}.
Here, the so called  \ms{}~\citep{gebser2019multi} solving is offered by \clingo{}. 

When implementing a \ms{} workflow, one inevitably has to choose how to influence the resulting answer sets.  
Facts, for once, permanently fix truth values of atoms, whereas values of external atoms are fixed but can be altered between solve calls. 
Furthermore, assumptions can be applied on regular atoms and in a more sophisticated manner, even truth values of (negated) clauses of atoms can be influenced during the solve call.
The different manipulation techniques possess their own advantages and disadvantages, requiring background knowledge for an informed decision.
A fitting choice depends on the requirements and characteristics of the applications as well as the setup.

This paper aims to show different approaches to reuse logic  programs on the showcase example of the popular and easy to grasp game \snakes{}. 
The objective of the game is to repeatedly finding a path from a snake to an apple on a grid.
\snakes{} has an iterative setting and can be encoded space efficient. 
Strategies to enforce winning the game include solving the  NP-hard~\citep{de2016complexity} problem of Hamiltonian Cycles. 
Due to matching complexity, this shows to be an interesting and suited example for an \asp{} implementation.
In the past, (\ms{}) \asp{} has been used to solve games such as Ricochet Robots \citep{Gebser_2015} and games in general  \citep{Thielscher2009}.

In this paper, we formalize the underlying problems and objectives to play \snakes{} and introduce two strategies to enforce winning the game. 
Then, we propose, explain, and analyze five approaches to implement one \snakes{} strategy. 
In an extensive empirical evaluation we compare differences among the approaches and share interesting insights to develop \ms{} applications. 
Our software is available online\footnote{https://github.com/elbo4u/asp-snake-ms} and includes a visually engaging image output of the game progression via \clingraph{}~\citep{hahn2022clingraph}.

\section{Preliminaries}
\subsection{Answer Set Programming}

A (disjunctive) program ${\Pi}$ in ASP is a set of rules $r$ of the form:
$$a_1;\dots{};a_m \;{:}{-}\; a_{m+1}, \dots{}, a_n, \;not \; a_{n+1},\dots{},\; not \; a_{o}.$$
where each \emph{atom}  $a_i$ is of the form: $p(t_1,\dots{},t_k)$, $p$ is a \emph{predicate} symbol of \emph{arity} $k$ and $t_1,\dots{},t_k$ are \emph{terms} built using constants and variables.  
For predicates with arity $k{=}0$, we will omit the parenthesis. 
A \emph{naf} (negation as failure) \emph{literal} is  of the form $a$ or $not\; a$ for an atom $a$. 
A rule is called \emph{fact} if $m{=}1$ and $o{=}0$, \emph{normal} if $m{=}1$ and \emph{integrity constraint} if $m{=}0$. 
Each rule can be split into a \emph{head}  $h(r)=\{a_1,\dots{},a_m\}$ and  a \emph{body} $B(r)=\{a_{m+1},\dots{},not\;a_o\}$, which divides into a positive part $B^+(r) = \{a_{m+1},\dots,a_n\}$ and a negative part $B^-(r) = \{a_{n+1},\dots,a_o\}$. 
An expression (i.e. term, atom, rule, program, ...) is said to be \emph{ground} if it does not contain variables. 
Let $M$ be a set of ground atoms, for a ground rule $r$ we say that $M \models r$ iff  $M \cap h(r) \neq \emptyset$ whenever $B^+(r)\subseteq M$ and $B^-(r)\cap M = \emptyset$. 
$M$  is a \emph{model} of $\mathcal{P}$ if $M \models r$ for each $r \in \mathcal{P}$. 
$M$ is a \emph{stable model} (also called \emph{answer set}) iff $M$ is a $\subseteq$-minimal model of the Gelfond-Lifschitz \emph{reduct} of 
$\mathcal{P}$ w.r.t. $M$. 
The reduct is defined as $\mathcal{P}^M = \{h(r) \gets{} B^+(r) |M \cap B^-(r) = \emptyset, r \in \mathcal{P} \}$~\citep{gelfond1988stable}.

To ease ASP programming, several language extensions such as  \emph{conditional literals},  \emph{cardinality constraints} and \emph{optimization statements} were introduced~\citep{CALIMERI2019}. 
Conditional literals are of the form $a{:}b_1,\dots{},b_m$ with naf-literals $a$ and $b_i$. 
They allow the conditional inclusion of atoms $a$ for condition $b_1,\dots{},b_m$ and  are especially useful in combination with variables. 
Cardinality constraints introduce compact counting of atoms with an upper bound $s$ and lower bound $t$. They are of the form $s\bl{}d_1;\dots{};d_n\br{}t$ with conditional literals $d_i$. 
Optimization statements of the form $\texttt{\#minimize\bl{}}w_1@p_1,t_1{:}c_1;\dots{};w_n@p_n,t_n{:}c_n\texttt{\br{}.}$ aim to minimize the sum over $w_i$ over a set of weighted tuples $(w_i,t_i)$ for priority level $p_i$ (default level $0$) under the condition $c_i$.
Additionally, the program \texttt{clingo} offers several extensions as well, such as ranges \texttt{l..u}, indicating any integer between \texttt{l} and \texttt{u}, including both. 

\subsection{Multi-Shot ASP in \clingo{}}
\clingo{} provides an API~\citep{gebser2019multi} to access the functionality of an \asp{} program within an imperative language such as python. 
The standard procedure is to create a \clingo{} control object in the wrapper program, assign one or more logic program inputs, ground it, and solve it. 
This setup allows to alter and repeatedly run logic programs (\emph{\ms{}}). 

Besides basic functionality, the \clingo{} control object provides access to several mechanics connected to the logic program.
Parametrized subprograms allow grounding of rules with values of constants determined at runtime, enabling flexible data management.
The solving is initialized via a solve call.
The corresponding function accepts attributes to influence the execution of the solving, such as \emph{assumptions}.
Assumptions are a list of atom/truth value pairs, which state valid mappings for the current solve call.
Another mechanic to influence truth values of atoms are \emph{externals}.
They are declared in the form $\texttt{\#external } a{:}b_1,\dots{},b_m \texttt{.}$ with conditional literal $a{:}b_1,\dots{},b_m$ within the logic program. 
An atom marked as external has a fixed truth value similar to a fact, except that in between solve calls the actual value can be changed via the function \textit{assign\_external}.
It is even possible to remove an external permanently by releasing it.

Next to assumptions and externals, there is also the possibility to directly influence the ongoing search for models.
During the solve call, truth values can be assigned to (negated) conjunctions of atoms (\emph{clauses} and \emph{nogoods}). 
In \clingo{} this functionality is accessible by either implementing a custom propagator or indirectly through the context of a model object.
The model object can be obtained by providing a callback model handling function to the solve call.

\subsection{Graphs, Paths, and Hamiltonian Cycles} %
A \emph{grid graph} $G(n,m)$ is an undirected graph formed by a rectangular grid of vertices~\citep{Itai1982}. 
The grid consists of $n$ rows and $m$ columns, denoted as $G(n,m) = (V, E)$, where $V$ represents the set of vertices with $|V|=n{\cdot}m$ and $E\subseteq V{\times}V$ represents the set of edges. 
Each vertex  $v(x,y)\in{}V$ corresponds to an intersection of a column $x$ and a row $y$, and each edge in $E$ connects adjacent vertices in the horizontal or vertical direction. 

A \emph{path} \mypath{} within a grid graph $(V, E)$ is a sequence of adjacent vertices $v_1, \dots, v_{n}$ denoted as list $[v_1, \dots, v_{n}]$, with $\mypath{}_i = v_i$ and $|\mypath{}|=n$, such that
 for every $i \in \{2, \ldots, n\}$ there exists an edge $(v_{i-1}, v_i) \in E$.
A subpath $\mypath{}'$ of \mypath{} covers a subsequence of vertices such that there exists a $c \in \mathbb{N}$, for which every $\mypath{}'_i$ equals $\mypath{}_{i{+}c}$.

A \emph{Hamiltonian cycle} (\hc{})~\citep{Itai1982} in a grid graph $G(n,m)=(V, E)$ is a path $\mypath{}$ of distinct vertices such that $(p_{n\cdot{}m}, p_1) \in E$. 
Therefore, an \hc{} is a cycle that passes through every vertex of the graph without repetition, following the adjacent edges of the grid structure. Only grid graphs with even numbers of vertices admit an \hc{}~\citep{Itai1982}.

\subsection{Snake Game}

\begin{figure}[!t]
	\centering
	\begin{subfigure}[b]{\figsize}
		\includegraphics[width=\textwidth]{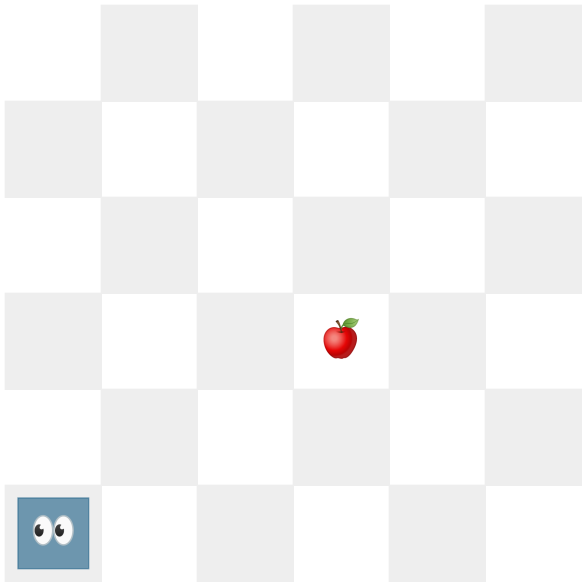}
		\caption{initial scenario }
		\label{fig:init:init}
	\end{subfigure}
	\hfill
	\begin{subfigure}[b]{\figsize}
		\includegraphics[width=\textwidth]{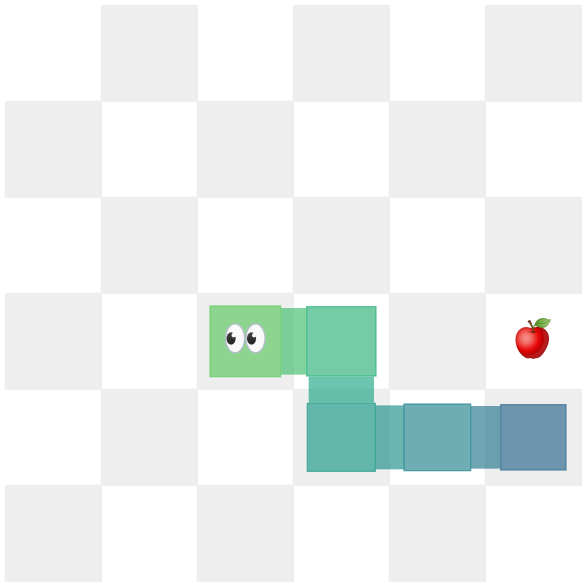}
		\caption{step - before}
		\label{fig:init:step1}
	\end{subfigure}
	\hfill
	\begin{subfigure}[b]{\figsize}
		\includegraphics[width=\textwidth]{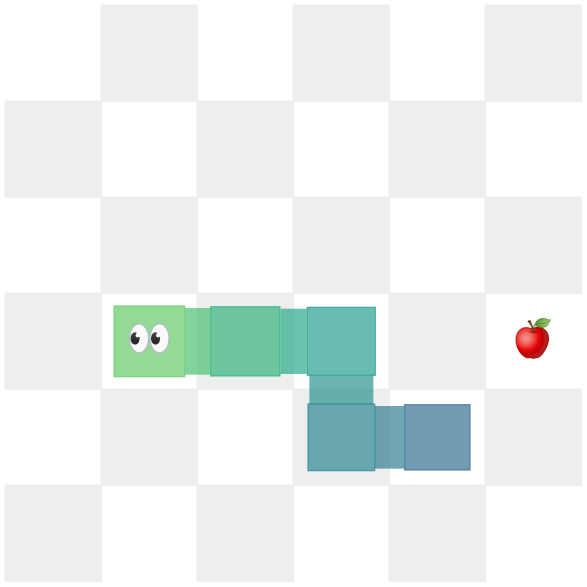}
		\caption{step - after}
		\label{fig:init:step2}
	\end{subfigure}
	
	\medskip
	\begin{subfigure}[b]{\figsize}
		\includegraphics[width=\textwidth]{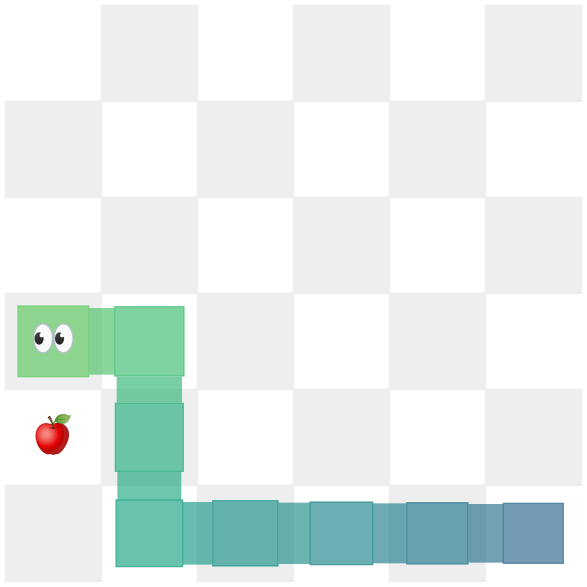}
		\caption{apple - before}
		\label{fig:init:apple1}
	\end{subfigure}
	\hfill
	\begin{subfigure}[b]{\figsize}
		\includegraphics[width=\textwidth]{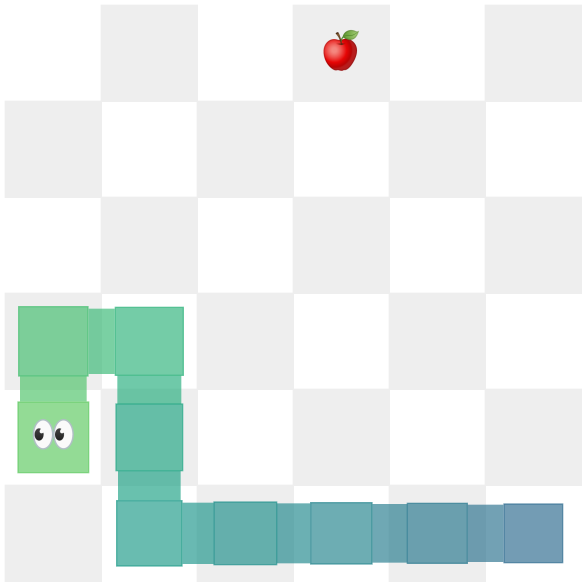}
		\caption{apple  - after}
		\label{fig:init:apple2}
	\end{subfigure}
	\hfill
	\begin{subfigure}[b]{\figsize}
		\includegraphics[width=\textwidth]{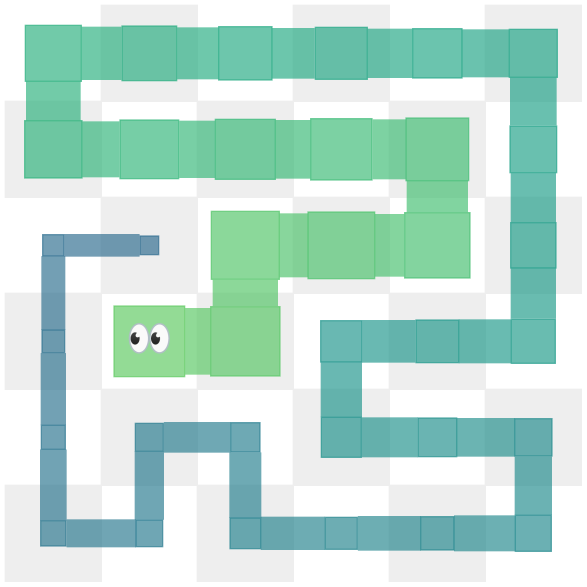}
		\caption{maximal length }
		\label{fig:init:complete}
	\end{subfigure}
	\medskip	
	\caption{Examples for a snake on a $6\times{}6$ grid generated by our software. The snake head is marked with eyes. %
	}
	\label{fig:intro}
\end{figure}

The game \snakes{} is a classic arcade game, where the user steers a snake's head on a grid similar to a grid graph of dimension $n{\times}m$. 
The snake $\snake{}$ can be interpreted as a list of grid coordinates, for example $\snake{} = [(6, 2), (5, 2), (4, 2), (4, 3), (3, 3)]$ for Figure~\ref{fig:init:step1} starting with the tail and ending with the head. 
Possible movements for the head are left, right, up and down to reach adjacent fields on the grid.
The snake body follows the head movement, meaning starting with $\snake{}$ after one \emph{step} the new snakes $\snake{}'$ body elements $\snake'{}_{i-1}$ now equal  $\snake{}_i$ as illustrated in Figures~\ref{fig:init:step1},\ref{fig:init:step2}.

A snake $\snake{}$ can increase its length by placing the head on a tile occupied by an apple with coordinates $\apple{}$, leading for the extended snake $\snake{}'$ to occupy the fields $[ \snake{}_{1}, \dots{}, \snake{}_{|\minisnake{}|}, \apple{}]$ as illustrated in Figures~\ref{fig:init:apple1},~\ref{fig:init:apple2}.
Now the apple is consumed and another apple appears on a random, unoccupied field.
The game starts with a snake of length one in the corner of the grid ($\snake = [(1,1)]$) and one apple on a random, unoccupied field as indicated in Figure~\ref{fig:init:init}. 
The maximal length of a snake is $n{\cdot}m$, since the snake fills all the fields and no apple can be placed, preventing any future growth of the snake as shown in Figure~\ref{fig:init:complete}. 
The game terminates if the head lands on a field of the body, leaves the grid, or the snake reached its maximal length. 
For the latter the game is considered won.

\section{Formalizing and Winning \Snakes{} Optimally with \asp{}}
We introduced the preliminaries and the game \snakes{}. We will now move towards defining the game objectives, then presenting strategies and their implementation in \asp{}.

\paragraph{Goal Specification.}

The game itself invokes the implicit goal to maximize the snake length before the game terminates. 
Our first objective is to guarantee winning the game. 
This means at every point the snake is able to reach maximal length, independent of apple placement (\goal{1}). 
Second, we aim to minimize the number of snake movements (\emph{steps}) to finish the game (\goal{2}). 
Third, we aim to minimize the computation time (\goal{3}).

\subsection{Problem Description for Iterations}

One \snakes{} game consists of $n{\cdot}m{-}2$ iterations of searching a path from a given snake \snake{} to a given apple \apple{} for given grid dimensions $n{\times}m$. 
Therefore, we can formalize a  problem description for one iteration.

\textbf{\gs{}.} 
\textit{
Given a path $\snake{}$, a goal vertex  $\apple{} \not \in \{\snake{}_1,  \cdots, \snake{}_{|\minisnake{}|}\} $ on a grid graph $G(n,m)$, derive a path  $\mypath{}=[\snake{}_1,  \cdots, \snake{}_{|\minisnake{}|},  \cdots, \apple{}]$, such that $p_a= \apple{}$ iff $a=|\mypath{}|$ and for every $1{\leq}j{<}k{<}|\mypath{}|$ with $\mypath{}_j = \mypath{}_k$, then $k-j\geq{}|\snake|$. 
}

\gs{} has three key points: the path has to start with \snake{}, the \apple{} vertex appears exactly once at the end and if a vertex repeats, enough steps lie between repetitions for the snake to move out of the way. \gs{} covers all possible solutions for one iteration of \snakes{}.

\begin{figure}[!t]
	\centering
	\begin{subfigure}[b]{\figsize}
		\includegraphics[width=\textwidth]{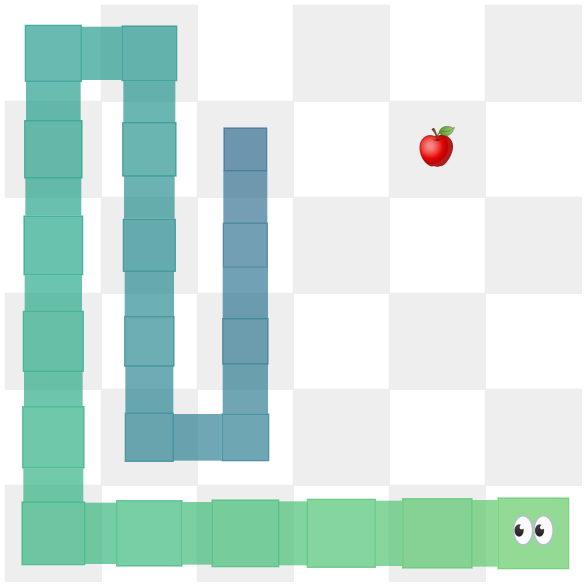}
		\caption{\hc{} input}
		\label{fig:hc:ex}
	\end{subfigure}
	\hfill
	\begin{subfigure}[b]{\figsize}
		\includegraphics[width=\textwidth]{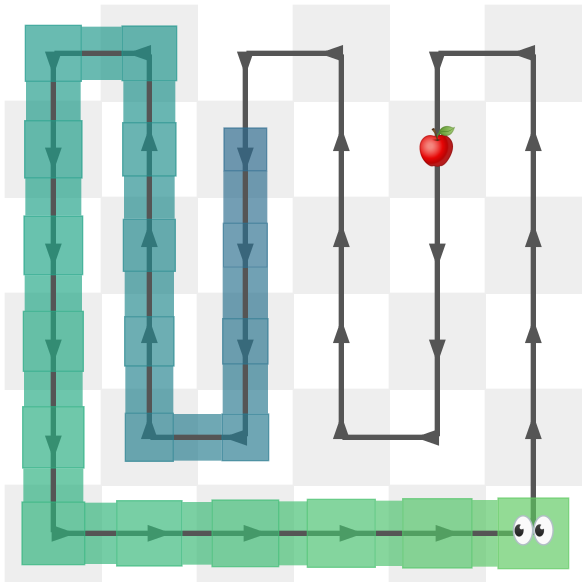}
		\caption{Generic \hc{}}
		\label{fig:hc:init}
	\end{subfigure}
	\hfill
	\begin{subfigure}[b]{\figsize}
		\includegraphics[width=\textwidth]{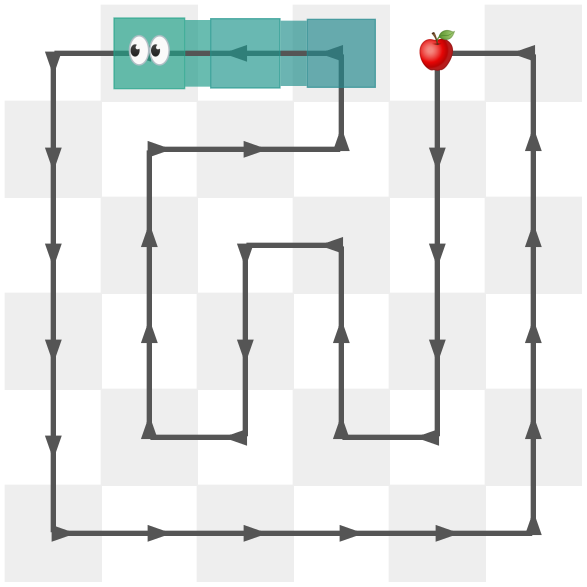}
		\caption{\mhs{} }
		\label{fig:hc:min}
	\end{subfigure}	
	\caption{Examples for a snake on a $6\times{}6$ grid. Black lines indicate \hc{}s. %
	}
	\label{fig:intro}
\end{figure}

While this problem description is sound to solve a single iteration, it does not ensure future iterations to be solvable (compare Figures~\ref{fig:init:apple1}, \ref{fig:init:apple2}). 
To force a game to be won (G1), each field has to be reachable in every iteration. %
An effective method to enforce reachability in all future iterations is deriving \hc{}s, since the snake will visit each field of the grid before reaching the field currently occupied by its tail.
Hence, we define two problems which include \hc{}s as well, where the second is an extension of the first.

\textbf{
	Hamiltonian Snake (\hs{}). }
\textit{
	Given a path $\snake{}$ for a grid graph $G(n,m)$, derive an \hc{} on $G$ starting with subpath \snake{}.
}

\textbf{
Minimal Hamiltonian Snake (\mhs{}). }
\textit{
Given a path $\snake{}$ and a goal vertex \apple{} with $\apple{} \not\in{} \{\snake{}_1,  \cdots, \snake{}_{|\minisnake{}|}\} $ for a grid graph $G(n,m)$, derive an \hc{} on $G$ starting with subpath $\snake{}$ which minimizes $i$ for $\mypath_i=\apple{}$. %
}

An example of the problem and its (minimal) Hamiltonian snake can be seen in Figures \ref{fig:hc:ex}, \ref{fig:hc:init}. 
Since \hc{}s can only be derived for grid graphs with even numbers of vertices~\citep{Itai1982}, we will handle grids with even number of fields only.

\paragraph{Complexity.}
\hs{} asks for a solution of an \np{} problem, putting in the complexity class of nondeterministic polynomial Function Problems  ($\fnp{}$). 
\mhs{} asks for a solution of an \np{} problem for which no other solutions with certain properties exist, putting \mhs{} in the complexity class of $\fnp{}$ with \np{} oracle calls  ($\fnp{}^{\np{}}$). 
We decided to define function problems, since strategies (see next paragraph) require actual paths as output.

\subsection{Snake Strategies}

To solve a whole snakes game, we will iteratively solve its  $n{\cdot}m{-}2$ \hs{} resp. \mhs{} problems (basic setup).
Starting in the first iteration with $\snake{}=[(1,1)]$, every iteration has a starting snake \snake{} and random apple coordinates $\apple \not\in\{\snake{}_1,  \ldots, \snake{}_{|\minisnake{}|}\} $ as variable input. 
From the output path $\mypath{} = [\snake{}_{1},\ldots{},\snake{}_{|\minisnake{}|}, \ldots, \mypath{}_{j},\ldots{},\mypath{}_{nm}]$ %
 with $\mypath{}_{j}=\apple{}$ we can derive the snake coordinates $\snake{}'$ for the next iteration:
$\snake{}' = [\mypath{}_{j-|\minisnake|}, \ldots{}, \mypath{}_{j}]$.     %

We will now introduce the \emph{naive} strategy, which aims to finish the game consistently by solving its \hs{} problems. 
In the first iteration, the naive strategy derives an \hc{}. 
This path is followed repeatedly, removing the requirement to generate new \hc{}s altogether. 
This strategy takes on average $\nicefrac{(nm-|\minisnake{}|+1)}{2}$ steps per iteration.
The first \hc{} can be derived in polynomial time, by replicating column two and three and row three from Figure \ref{fig:hc:init} for arbitrary grid sizes\footnote{An \hc{} can only be derived for grid graphs with even number of fields, therefore the grid is required to  have at least one even dimension.}.
The number of steps per iteration is bound by the number of fields, consequently putting the naive strategy in polynomial time as well. %

The naive strategy aims playing safe (\goal{1}). 
To minimize the total number of steps (\goal{2}) we introduce the  \emph{conservative} strategy, which allows the \hc{} to change in between iterations.
Hence, we follow the basic setup and solve \mhs{} problems. 
Solving \mhs{} lies in $\fnp{}^{\np{}}$, which places the conservative strategy in the same complexity class.

Addressing the computation time (\goal{3}) we have to discuss and compare different implementations and design choices, which will be covered throughout this publication.

\subsection{Multi-Shot Implementation Approaches}

By following the conservative strategy, playing snakes boils down to generating \hc{}s efficiently. 
\asp{} is well suited for this task, \np{}-Problems can be easily expressed. 
Furthermore,  \snakes{} can be implemented quite space efficient as well: Algorithm~\ref{alg:lp} shows the base logic program for the initial iteration, with \texttt{head/1} and \texttt{apple/1} as variable input. 
\texttt{field/1} defines the nodes of the grid, \texttt{connected/2} the edges,  
\texttt{next/2} represent the path choices and \texttt{path/1} ensures a closed cycle. 
Finally, with \texttt{mark/1} the path segment between head and apple is marked to be minimized in the last line.
In latter iterations, a snake body can be enforced by manipulating corresponding \texttt{next/2} atoms. 

\begin{algorithm}[t]
	\caption{base logic program to solve one iteration of snakes}
	\label{alg:lp}
	\KwIn{grid dimension $\texttt{n}\times{}\texttt{m}$, position of head (\texttt{head/1}) and apple (\texttt{apple/1})}
	{\ttfamily 
		
		field(({\cvar X},{\cvar Y})) :- {\cvar X}{\cmat =}\cnum{1}\cmat{..}n, {\cvar Y}{\cmat =}{\cnum 1}\cmat{..}m.\;
		connected((\cvar{X},\cvar{Y1}),(\cvar{X},\cvar{Y2})) :- \cmat{\string|}\cvar{Y1}\cmat{-}\cvar{Y2}\cmat{\string|=}\cnum{1}, field((\cvar{X},\cvar{Y1})), field((\cvar{X},\cvar{Y2})).\;
		connected((\cvar{X1},\cvar{Y}),(\cvar{X2},\cvar{Y})) :- \cmat{\string|}\cvar{X1}\cmat{-}\cvar{X2}\cmat{\string|=}\cnum{1}, field((\cvar{X1},\cvar{Y})), field((\cvar{X2},\cvar{Y})).\;
		\vspace{0.5\baselineskip} 
		
		\cnum{1} \string{ next(\cvar{XY},\cvar{XY\textquotesingle}) : field(\cvar{XY} ), connected(\cvar{XY},\cvar{XY\textquotesingle}) \string}  \cnum{1} :- field(\cvar{XY\textquotesingle}).\;
		\cnum{1} \string{ next(\cvar{XY},\cvar{XY\textquotesingle}) : field(\cvar{XY\textquotesingle}), connected(\cvar{XY},\cvar{XY\textquotesingle}) \string} \cnum{1} :- field(\cvar{XY} ).\;
		\vspace{0.5\baselineskip} 
		
		path(\cvar{XY}) :- field(\cvar{XY}), head(\cvar{XY}).  \;%
		path(\cvar{Next}) :- path(\cvar{XY}), next(\cvar{XY},\cvar{Next}). \;
		:- field(\cvar{XY}), \cmat{not} path(\cvar{XY}).  \;
		\vspace{0.5\baselineskip} 
		
		mark(\cvar{XY}) :- field(\cvar{XY}), head(\cvar{XY}). \;%
		mark(\cvar{Next}) :- mark(\cvar{XY}), next(\cvar{XY},\cvar{Next}), \cmat{not} apple(\cvar{XY}). \;
		\cnum{\#minimize}\string{ \cnum{1},\cvar{XY} : mark(\cvar{XY}) \string}.
	}
\end{algorithm}

\asp{} base Algorithm \ref{alg:lp} can be utilized to iteratively solve snake in different ways. %
A simple approach is to ground and solve the logical program for each iteration and discard it afterwards ({\aos{}}).
However, there are several reasons to reuse a logic program. 
For instance, time intense grounding or minimal changes can be exploited, especially for iterative progressing problems. 
There are several approaches to reuse a logic program. 
Main contribution of this work is a compilation and demonstration of different approaches.
Therefore, we will now introduce five different \clingo{} implementations to solve \snakes{}. 
Technically, they will mainly differ in the method to manipulate \texttt{next/2} atoms.

An outline of the wrapper functionality for \ms{} approaches can be seen in Algorithm \ref{alg:base}, which implements the conservative strategy.
In Lines 1-3 the logic program, \snake{} and step counter are initialized. 
The initialization (Line 1) for \ms{} approaches includes grounding \texttt{apple/1} and \texttt{head/1} as externals:  \\
\texttt{
\hphantom{....}\#external apple(XY): field(XY). \\
\hphantom{....}\#external head(XY) : field(XY).}\\ 
Without these (external) atoms, the optimization statement can not be formulated.
In a loop, \apple{} is picked at random (Line 5) and corresponding external values for the apple position and snake head are set (Lines 6, 7, 9, 10). 
The main part is represented by he  $\textit{retrieve}$ function (Line 8). 
Here, the values of \snake{} will be tied to corresponding \texttt{next/2} atoms from the logic program. 
Furthermore, the solve process is started and resulting models are managed and converted into paths. 
The implementation of the \textit{retrieve} function depends on the approach (see Algorithms \ref{alg:aos}-\ref{alg:nogood2}). 
In Line 11, the new snake position is derived from the path. 
The for loop stops once \snake{} reached maximal length or no path could be derived.

\begin{algorithm}[t]
	\caption{Main Algorithm for iteratively computing snake paths}
	\label{alg:base}
	\KwIn{grid dimension $n\times{}m$}
	\KwOut{number of total steps}
	$\Pi \gets \textit{init}(n,m)$\Comment*[r]{ground Algorithm 1; apple \& head as external} 
	$\snake{} \gets [(1,1)]$\Comment*[r]{\snake{} starts at position (1,1) with |\snake{}|=1}
	$\textit{steps} \gets 0$\Comment*[r]{step counter}
	\Do{$|\snake{}|<n\cdot{}m$ \KwSty{and} \KwSty{not} $path = \{\}$}{
		$\apple{} \gets \textit{generate\_apple}((n,m), \snake{})$\Comment*[r]{place on random, non-\snake{} field} %
		
		$\Pi \gets \textit{set\_external}(\Pi,\texttt{apple(}\apple{}\texttt{)}, \textit{True})$\Comment*[r]{set external atoms} %
		$\Pi \gets \textit{set\_external}(\Pi,\texttt{head(}\snake{}_{|\minisnake{}|}\texttt{)}, \textit{True})$\;
		$\Pi,\textit{path} \gets \textit{\underline{retrieve}}(\Pi,\snake{})$\Comment*[r]{enforce \snake{} onto next/2, generate path} %
		$\Pi \gets \textit{set\_external}(\Pi,\texttt{apple(}\apple{}\texttt{)}, \textit{False})$\Comment*[r]{undo external atoms} %
		$\Pi \gets \textit{set\_external}(\Pi,\texttt{head(}\snake{}_{|\minisnake{}|}\texttt{)}, \textit{False})$\;
		$\snake{}, s \gets \textit{follow\_path}(\snake{},\textit{path}, \apple{})$ \Comment*[r]{\snake{} follows ${path}$ until \apple{}, $s$ steps}
		
		$\textit{steps} \gets \textit{steps} + \textit{s}$ \Comment*[r]{update step count} %
	}
	$\KwSty{return}\; \textit{steps}$\;
\end{algorithm}

\paragraph{\Aos{}.}
\aos{} can be understood as the default approach for logic programs.
Here, the current position of the snake and the apple are introduced as facts, resulting in permanently fixing the current values.
Each iteration requires grounding and solving a new logic program from scratch.
While \aos{} sounds rather wasteful (especially in an iterative setting), it comes with clear advantages as well: 
There is no need to consider mechanics of extending or manipulating logic programs, thus creating a simple, stable, straight forward and easy to debug solution. 
Therefore \aos{} is well suited for fast development projects with easy code maintenance or noncritical execution times.

Implementing \aos{} uses a similar outline as Algorithm \ref{alg:base}, with small changes.
Line 1 would be moved into the loop before the apple placement. 
Instead of setting externals (Lines 6, 7, 9, 10), facts are added to the logic program before solving: 

$\Pi \gets \Pi{}\cup{}\textit{ground}(\{\texttt{apple(}\apple{}\texttt{).}\; \texttt{head(}\snake{}_{|\minisnake{}|}\texttt{).}\})$.
 
The \textit{retrieve} function for \aos{} is outlined in Algorithm \ref{alg:aos}.
The function translates \snake{} into equivalent \texttt{next} atom facts to be added to $\Pi{}$.
The program is subsequently solved and the path derived from the model returned.
To fit the interface of the \ms{} algorithm, the program $\Pi{}$ is returned as well.

\begin{algorithm}[t]
	\caption{\textit{retrieve} for \aos{} }
	\label{alg:aos}
	\KwIn{program $\Pi$, snake position list \snake{}}
	\KwOut{program $\Pi$, path}
	\For{$ i = 1\, ..\, |\snake{}|{-}1$}{
		$\Pi \gets \Pi \cup \textit{ground}(\{\texttt{next(}\snake{}_{i}\texttt{,}\snake{}_{i{+}1}\texttt{).}\}$) \Comment*[r]{transfer \snake{} values to next/2}  
	}
	$\textit{model} \gets \textit{solve}(\Pi)$ \Comment*[r]{generate model} 
	$\KwSty{return}\; \Pi, \textit{extract\_path}(\textit{model})$ \Comment*[r]{extract path out of model} 
\end{algorithm}

\paragraph{\Redo{}.}

As for \ms{} implementations, adding and removing rules \citep{gebser2019multi, Gebser_2015} will be called the \redo{} variant.
Adding rules is comparatively easy, as a program is essentially a set of rules. 
To disable or remove rules, we utilize external atoms, since their truth values are assigned outside of the logic program.
Basically, an external atom is introduced as guard to a temporary rule.
Once the external guard is released, the connected rules are released as well. 
Released externals can not be reused. 

We will explain the mechanics through the implementation of the \textit{retrieve} function from Algorithm \ref{alg:redo}, fitting into Algorithm \ref{alg:base}.
In our case an atom of \texttt{step/1} is declared to be external (Line 1).
Now constraints enforcing the current \snake{} positions onto \texttt{next/2} are added with the external as guard (Line 3).
The external is set to \textit{True} (Line 4) and the program solved. 
To remove the current restrictions on $\Pi{}$, the external is released afterwards (Line 6) and $\Pi{}$ is suited to be used in another iteration.

In current literature, usually \redo{} is utilized, since it allows adding of previously unspecified rules and atoms.
This comes in handy, especially for logic programs with rising horizons. 
Therefore \redo{} (along with \aos{}) is the most flexible of all approaches. 
On the downside, expanding a ground logic program may require a deeper understanding of its underlying dependencies, as existing rules need to be compliant with the added rules.
Reducing the number of temporary rules (to streamline the solving process) poses another challenge, since its realization usually adds complexity to maintain the code.

\begin{algorithm}[t]
	\caption{\textit{retrieve} for \redo{}; \, Input: $\Pi$, \snake{}; \, Output: $\Pi$, \textit{path}}
	\label{alg:redo}
	
	$\Pi \gets \Pi \cup \textit{ground}(\{\texttt{\#external step(}|\snake{}|{\texttt{).}}\})$ \Comment*[r]{introducing the guard} 
	\For{$ i = 1\, ..\, |\snake{}|{-}1$}{
		$\Pi \gets \Pi \cup \textit{ground}(\{\texttt{:- }\; \texttt{step(}|\snake{}|\texttt{),} \texttt{ \cmat{not} next(}\snake{}_{i}\texttt{,}\snake{}_{i{+}1}\texttt{).}\}$) \Comment*[r]{adding rules} 
	}
	$\Pi \gets \textit{set\_external}(\Pi, \texttt{step(}|\snake{}|\texttt{)}, \textit{True})$ \Comment*[r]{activating guard} 
	$\textit{model} \gets \textit{solve}(\Pi)$\;
	$\Pi \gets \textit{release\_external}(\Pi, \texttt{step(}|\snake{}|\texttt{)})$ \Comment*[r]{removing guard and its rules} 
	$\Pi \gets \textit{cleanup}(\Pi{})$ \Comment*[r]{cleanup after release} 
	$\KwSty{return}\; \Pi, \textit{extract\_path}(\textit{model})$ \Comment*[r]{generate path out of model} 
\end{algorithm}

\paragraph{\Preground{}.}

In a restricted setting, adding and removing rules might not be a required feature and would result in unnecessary computational overhead. 
Pregrounding all required temporary rules might improve computation times, reduce complexity of the code and introduces a neat interface between the wrapper program and the logic program. 
Each temporary rule acquires at least one external predicate in the body to be switched on and off again. 
Pregrounding does not include the option to add rules on the fly since all possible changes are introduced once in the beginning. %
Additionally, a vast amount of inactive rules might slow down the program as well. 
However, \preground{} might perform well on compact programs with lots of iterations, since in comparison to \redo{} no overhead is generated.

Algorithm \ref{alg:base} can be extended for preground by adding the following rules to the initial grounding in Line 1: \\
$\texttt{\#external prenext(\cvar{X},\cvar{Y}) : connected(\cvar{X},\cvar{Y}).}$  \\
$\texttt{:- prenext(\cvar{X},\cvar{Y}), \cmat{not} next(\cvar{X},\cvar{Y}), connected(\cvar{X},\cvar{Y}).}$ \\
By setting the external \texttt{prenext/2} to \textit{True}, we can enforce the corresponding \texttt{next/2} atom to be  \textit{True} as well.
Algorithm \ref{alg:preground} indicates the functionality of the \textit{retrieve} function, which boils down to translating the snake into corresponding \texttt{prenext/2} atoms and temporary activating them before solving. 

\begin{algorithm}[t]
	\caption{\textit{retrieve} for \preground{}; \, Input: $\Pi$, \snake{}; \, Output: $\Pi$, \textit{path}}
	\label{alg:preground}
	
	\For{$ i = 1 .. |\snake{}|-1$}{
		$\Pi \gets \textit{set\_external}(\Pi, \texttt{prenext(}\snake{}_i\texttt{,}\snake{}_{i{+}1}\texttt{)}, \textit{True})$  \Comment*[r]{enforce \snake{} positions}  
	}
	$\textit{model} \gets \textit{solve}(\Pi)$\;
	\For{$ i = 1 .. |\snake{}|-1$}{
		$\Pi \gets \textit{set\_external}(\Pi, \texttt{prenext(}\snake{}_i\texttt{,}\snake{}_{i{+}1}\texttt{)}, \textit{False})$\Comment*[r]{deactivate}  
	}
	$\KwSty{return}\; \Pi, \textit{extract\_path}(\textit{model})$\;
\end{algorithm}

\paragraph{\Assume{}.}
Instead of handling externals and altering the logic program every time, we can also fix truth values for non-external atoms via assumptions. 
This option is limited to truth values of atoms, therefore in many cases the introduction of auxiliary rules and atoms is advised, comparable to \preground{}.
Assumption atoms do not require to be declared as such and they are provided as a list of atom/truth value assignments to the solve call.
In comparison to externals, assumptions are not persistent and have to be formulated for every solve call, which may lead to a less organized setup in comparison to \preground{}.
The naming convention should also be considered: names of externals are usually picked for easy access in the wrapper program, whereas assumption require to operate on predefined atom names. 
Handling  atoms with complex terms in \clingo{} can become convoluted quite fast, decreasing the readability of the code. 
\assume{} and \preground{} compare in functionality, with \preground{} contributing a simpler interface and  \assume{} being more flexible.

The \textit{retrieve} function for the \assume{} approach is shown in  Algorithm \ref{alg:assume}. 
In \clingo{} assumptions are provided as an attribute to the solve call.
Enforcing the \snake{} position via assumptions does not require any alterations of the logic program.

\begin{algorithm}[t]
	\caption{retrieve for \assume{}; \, Input: $\Pi$, \snake{}; \, Output: $\Pi$, \textit{path}}
	\label{alg:assume}
	$\textit{assume} \gets [\,]$ \Comment*[r]{start with empty list of assumptions} 
	\For{$ i = 1 .. |\snake{}|-1$}{
		$\textit{assume.append}( (\texttt{next(}\snake{}_i\texttt{,}\snake{}_{i{+}1}\texttt{)}, \textit{True}))$ \Comment*[r]{add atom/value tuple} 
	}
	$\textit{model} \gets \textit{solve}(\Pi, \textit{assumption} = \textit{{assume})}$\;%
	$\KwSty{return}\; \Pi, \textit{extract\_path}(\textit{model})$\;
\end{algorithm}
\paragraph{\Nogood{}.}
Next to assignments of truth values to (external) atoms, \clingo{} also offers mechanics to influence the search progress.
During the search, proven sub-results are stored in form of (negated) conjunctions of literals, which are called \emph{clauses} and \emph{nogoods}\citep{weinzierl2017blending}.
Adding custom nogoods to an ongoing search can be utilized as \ms{} approach.
Comparable to assumptions, we are able to assign truth values to not only atoms but clauses during a solve call.
The \nogood{} approach therefore can handle rules similar to \preground{} without prior definition, since truth values can be assigned to arbitrary conjunctions of literals. 
In \clingo{} nogoods require sophisticated knowledge to access: either via a customized propagator or via accessing a model from the solve call.
We implemented the latter, which causes the issue of adding constraints after obtaining the initial model. 
Like this, the first model does not respect the snake placement. 
This can result in an unobtainable short path, which renders the subsequent optimization futile.
However, the system can be steered to generate a predefined model using heuristics as described in the next paragraph.

To manipulate nogoods, a function to process obtained models needs to be assigned in the solve call (here: \textit{my\_func}). 
The \textit{retrieve} function boils down to assigning a function for model handling as shown in Algorithm \ref{alg:nogood1}. 
\begin{algorithm}[t]
	\caption{retrieve for \nogood{}; \, Input: $\Pi$, \snake{}; \, Output: $\Pi$, \textit{path}}
	\label{alg:nogood1}
$model \gets{} solve(\Pi, \text{on\_model} = \textit{my\_func})$\; 
$\KwSty{return}\; \Pi, \textit{extract\_path}(\textit{model})$ 
\end{algorithm}
The function \textit{my\_func} has access to the model object during the solve call. 
It is its only argument. 
Additional objects can be provided via object oriented programming \citep{Gebser_2015}. 
A partial implementation of \textit{my\_func} can be seen in Algorithm \ref{alg:nogood2}. 
At first, the function checks if the current model is the first obtained model and compliant with our not yet enforced restrictions.
In our implementation, this model is marked with a special atom (\texttt{dummy}). 
If this is the case, a clause is added to enforce the current snake placement.

\begin{algorithm}[t]
\caption{\textit{my\_func} - code snipped to add \snake{} as search restriction}
\label{alg:nogood2}
\KwIn{model $model$}
$\cdots{}$\;	
\If{$\texttt{\text{dummy}} \in \textit{model}$
}{	
\For{$ i = 1 .. |\snake{}|-1$}{
		$\textit{model.context.add\_clause}(\texttt{next(}\snake{}_i\texttt{,}\snake{}_{i{+}1}\texttt{)}, \textit{True})$
}}
$\cdots{}$
\end{algorithm}

To summarize: 
The \nogood{} approach combines the possible expressivity of \preground{} without the baggage of inactive rules with the flexibility of assumptions. 
However, access to nogoods requires sophisticated knowledge of the \clingo{} API. 
If the nogoods are set via the model object, mechanics to enforce a predefined first model might be necessary.

\paragraph{Initial Model via Heuristics.}
For the introduced strategies, we  can generate one \hc{} from the previous iteration or a generic \hc{} for the first iteration (compare Figure \ref{fig:hc:ex}).
Therefore, we can jump-start the current solve process, guaranteeing at least one model for the optimization in the conservative strategy. %
We can utilize heuristics \citep{PotasscoUserGuide19} to inject this dummy model  within the logic program in \clingo{}:\\
\hphantom{......}\texttt{\string{dummy\string}.}\\
\hphantom{......}\texttt{\#heuristic dummy. [ \cnum{99}, true]}\\
\hphantom{......}\texttt{\#external heur(\cvar{X},\cvar{Y}) : connected(\cvar{X},\cvar{Y}).}\\
\hphantom{......}\texttt{:- dummy, heur(\cvar{X},\cvar{Y}), \cmat{not} next(\cvar{X},\cvar{Y}).}\\
For this, an independent \texttt{dummy} atom is introduced  and marked with a high preference to be set to \textit{True}.
Additional external atoms (i.e. $\texttt{heur/2}$) can be preset to influence the actual effect of the $\texttt{dummy}$ atom to the models where it is contained.
We are now able to set a dummy model, which benefits computation times and is required to implement the \nogood{} approach via model object.

\section{Experimental Evaluation}

\paragraph{\textbf{Setup.}}

To compare performance of the different approaches, we implemented all five approaches of the conservative strategy and let them solve 100 snake games for different square grid sizes ($n=m, n\in\{6,8,10,12,14,16\}$).
The experiments run on a MacBook Pro (2017, 16~GB RAM, Intel Core i7, 2.8~GHz), with clingo v. 5.4.0 and python v. 3.7.4. 
For each run, each of the $n\cdot{}m{-}2$ iterations has a 60 seconds timeout for the solve call.
Image generation via \clingraph{}~\citep{hahn2022clingraph} is disabled. 

We utilize a symmetry breaking method to mirror the grid, such that the head always lies in the first quadrant of the grid. 
We used \asp{}-Chef \citep{alviano2023Chef} for prototyping of the logic program. 
Our implementation and logfiles are available online\footnote{software: https://github.com/elbo4u/asp-snake-ms, logfiles: doi.org/10.5281/zenodo.13234723}.

\begin{table}
	\centering
	\begin{tabular}{llrrrrrr}
		\toprule
		&$n\times{}m $ &$6\times{}6 $  & $8\times{}8 $  &  $10\times{}10 $ & $12\times{}12 $ & $14\times{}14 $& $16\times{}16 $\\
		\midrule
		&\aos{} &    213 &  576 &  1235 &  2441 &  \underline{5519} &  \underline{10157} \\
		&\redo{} &    208 &  572 &  1226 &  2411 &  4582 &   7445 \\
		(a)	 &\preground{} &  216 &  563 &  1236 &  2374 &  4540 &   7482 \\
		&\assume{} &    210 &  563 &  1234 &  2396 &  4508 &   7540 \\
		&\nogood{} &    212 &  559 &  1240 &  2428 &  4580 &   7523 \\
		\midrule
		
		& \aos{} &             			0.159 &\textbf{2.28}&\textbf{71.83}&     	621 &    	2216 &    4359 \\
		& \redo{} &             			0.066 &        3.42 &        90.07 &     	674 &    	1966 &    3869 \\
		(b)	& \preground{} &             		0.060 &        3.32 &        94.71 &\textbf{620}&    	1978 &    3870 \\
		& \assume{} &             \textbf{	0.059}&        4.78 &        97.48 &     	628 &\textbf{1944}&   \textbf{3853} \\
		& \nogood{} &           			0.061 &        3.16 &        94.70 &     	702 &    	1951 &    3877 \\
		\bottomrule
	\end{tabular}
	\caption{(a) Average number of total steps per run. Outliers are underlined.\\
		(b) Average total time per game in seconds, timeout $60s$, best value bold.}
	\label{tab}
\end{table}
\paragraph{\textbf{Evaluation.}}

We expect \snakes{} to be feasible  to a certain extend (\E{1}).
Furthermore, we expect \ms{} and \aos{} to differ performance wise (\E{2}) and we expect \ms{} to outperform \os{} due to the grounding bottleneck (\E{3}). As educated guess, we expect \redo{} to outperform \preground{} based on unused externals (\E{4}). Due to implementation details, we expect \assume{} to terminate slightly faster than \nogood{} (\E{5}).

According to our three objectives, we are interested in three features: the win/loose ratio (\goal{1}), the number of total steps  (\goal{2}) and the total time  (\goal{3}). 
Since our strategy uses the previous \hc{} as starting model, there is a $100\%$ win ratio and \goal{1} is fully met. 
For \goal{2}, we compare the total number of steps, which are listed in Table \ref{tab}(a). 
For most grid sizes the numbers do not vary significantly. 
This is not surprising, since they implement the same algorithm. 
However \aos{} falls behind for larger grid sizes %
 ($n\geq{}14$, \E{2}). 

As for the total time (\goal{3}, Table \ref{tab}(b)), the \ms{} approaches do not differ significantly except for \aos{} (\E{2}). 
For  grid $6{\times}6$, the grounding impacts the total time negatively, placing \aos{} last (\E{3}). 
This changes for grid sizes $8{\times}8$ and $10{\times}10$, where \aos{} consistently leads in total time.
For $12{\times}12$ time consumption is on par and for larger grids ($14{\times}14$, $16{\times}16$) \aos{} has the highest total time. 

\begin{figure}
	\centering
	\includegraphics[width=1\textwidth]{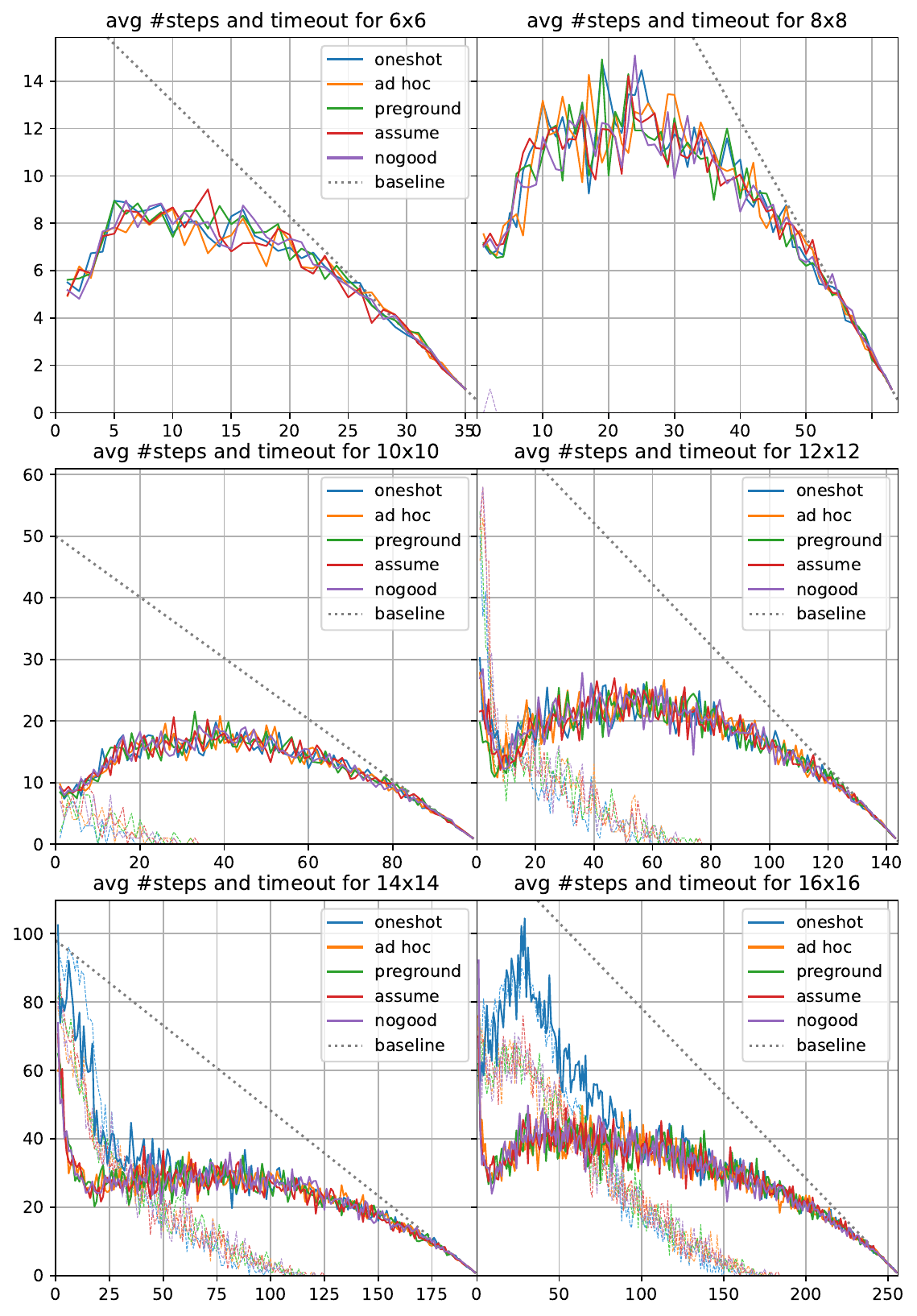}
	\caption{Average number of steps (solid lines) and timeout rate in percent (dashed lines) per iteration for different grid sizes for timeout 60s. Dotted gray line is the expected step count for the naive strategy (baseline).}
	\label{fig:steps}
\end{figure}

The data hint a disadvantage for \aos{} for larger grids, therefore we will analyze the step count per iteration as seen in Figure \ref{fig:steps}. 
For each grid size, the average number of steps to reach the apple is plotted against the iteration number, which equals the current length of the snake. 
The expected step count for the naive strategy is hinted as gray dotted line. 
For smaller grids (${<}12{\times}12$) all implementations perform on par and the curves show roughly the same pattern.
For grid $10{\times}10$ the curve pattern changes.
This can be explained with the timeout ratio (colored dashed line), meaning all approaches would have required more time to finish the optimization in the first iterations. %
The timeout was triggered for grid $8{\times}8$ as well, but seemed to mainly affect the last step of confirming the optimum, while for $10{\times}10$ the timeout took place during the active optimization phase (\E{1}). 
For grid $12{\times}12$ the curves for the different approaches do not differ much. 
However, for grids $14{\times}14$ and $16{\times}16$ the curve for \aos{} differs  significantly from the \ms{} approaches:
During the first iteration, all \ms{} approaches face the same timeout rate and impact as \aos{}.
However, their performance improves substantially faster in the following iterations.
We assume the \ms{} advantage is based on utilizing past search progress in form of learned nogoods and seems to increase for larger grid sizes.

The different \ms{} approaches operate on similar levels. %
\assume{} performed best in terms of computation times. 
However, the differences are minor and vary for different setups.
As educated guess, we would  expect an advantage of \redo{}  over \preground{} for high amounts of temporary rules. 
Our \preground{} implementation introduces a comparatively small number of temporary rules and the data do not support a disadvantage of \preground for this implementation (\E{4}). 
Also, we expect \assume{} to outperform \nogood{}, since the handling of the nogoods counts towards the solving time. 
This advantage is probably too minor to manifest in the data (\E{5}).

\paragraph{\textbf{Summary.}}
All approaches guarantee winning snakes.
\aos{} dominated the comparison, until resources were limited by the timeout.
The other approaches struggled with the timeout as well, but could utilize the previous solving history, leading to a reduction of timeouts through the iterations. 
The \ms{} approaches performed on similar levels, individual rankings were not consistent throughout different setups.

\subsection*{Related Work}
\hc{}s for grid graphs are well studied \citep{Itai1982}.
 \asp{} was utilized  to find \hc{}s in general graphs \citep{jeliahc, encodingselhc}.
Several implementations for AI approaches on playing \snakes{} are available. 
A variety of search algorithms is used to play \snakes{}, varying from basic strategies such as A* and \hc{}s \citep{snakeBachelorAppaji2020}\footnote{ https://johnflux.com/2015/05/02/nokia-6110-part-3-algorithms/} \footnote{ https://github.com/BrianHaidet/AlphaPhoenix/} to more sophisticated approaches \citep{snakeEvoYeh2016,Wei2018}.
To our knowledge no further  \snakes{} implementation in logic programming has been proposed. 
However, \ms{} \asp{} has been used to solve iterative games such as Towers of Hanoi~\citep{gebser2019multi} and Ricochet Robots~\citep{Gebser_2015}. Several (iterative) games are already implemented in \asp{}, such as Rush Hour~\citep{cian2022modeling}, Sokoban~\citep{incrementalASP} and Icosoku~\citep{rizzo20223cosoku}. 

Other solvers such as DLV2 support \ms{} as well~\citep{dlv2ms}.

\section{Conclusion and Future Work}

This paper aims to compile different \asp{} \ms{} approaches on an compact and hands-on showcase. 
While  \aos{} is straightforward to implement, applications with limited resources (such as timeouts) may substantially benefit from \ms{} implementations, as demonstrated in our evaluation. 
For our example, the grounding bottleneck was not an issue due to a minimalistic implementation.
\aos{} outperformed the \ms{} approaches as long as the timeout was not met. 
For harder problems, the timeout had a huge impact on the performance. 
The \ms{} approaches could utilize the search progress from previous solving attempts, reducing the impact of the timeout in the following iterations and therefore outperforming \aos{}.
The different \ms{} approaches possess varying characteristics, therefore the optimal choice depends on the application and the setup. 
The approaches \redo{} and \assume{} have indicated stable performance and are comparatively easy to implement.

Future work might entail another showcase application to demonstrate performance differences between the different \ms{} approaches. 
In addition, while the definition of \mhs{} is sound, a different problem description may lead to even shorter paths while still guaranteeing winning the game. An example scenario can be seen in Figure \ref{fig:hc:min}, where shorter paths to enforcing a win can be derived. 
For this more sophisticated problem only the snake end positioning has to be able to form an \hc{}. 
However, this new problem, resulting strategies, extended implementations and additional evaluation go beyond the scope of this paper.
Also, the introduced strategies rely on a continuous setting, meaning our strategies can not be started from an arbitrary snake placement as seen in Figure \ref{fig:init:apple1}.
Strategies starting with random snake placements may be covered in the future.

\paragraph*{Acknowledgments} 
The authors are stated in alphabetic order. This work was supported by BMBF in project { 01IS20056\textunderscore{}NAVAS} and in BMBF grant ITEA-01IS21084 ({InnoSale}),  DFG grant 389792660 ({TRR 248}), and in DAAD grant 57616814 ({SECAI}).

\bibliographystyle{tlplike}
\bibliography{snakesplus}

\end{document}